\def\year{2020}\relax
\documentclass[letterpaper]{article} % DO NOT CHANGE THIS
\usepackage{aaai20}  % DO NOT CHANGE THIS
\usepackage{times}  % DO NOT CHANGE THIS
\usepackage{helvet} % DO NOT CHANGE THIS
\usepackage{courier}  % DO NOT CHANGE THIS
\usepackage[hyphens]{url}  % DO NOT CHANGE THIS
\usepackage{graphicx} % DO NOT CHANGE THIS
\usepackage{graphicx}  % DO NOT CHANGE THIS
\usepackage[utf8]{inputenc} % allow utf-8 input
\usepackage[T1]{fontenc}    % use 8-bit T1 fonts
\usepackage{hyperref}       % hyperlinks
\usepackage{url}            % simple URL typesetting
\usepackage{booktabs}       % professional-quality tables
\usepackage{amsfonts}       % blackboard math symbols
\usepackage{nicefrac}       % compact symbols for 1/2, etc.
\usepackage{microtype}      % microtypography
\usepackage{lipsum}
\usepackage{threeparttable}
\usepackage{multicol}
\usepackage{multirow}
\usepackage{setspace}
\usepackage{pdfpages}
\usepackage{algpseudocode}  
\usepackage{algorithm}  
\usepackage{amsmath}  
\usepackage{float}
\usepackage{subfigure} 
\usepackage{array}

\title{Complementary Fusion of Multi-Features and Multi-Modalities in Sentiment Analysis}
\author{
	Feiyang Chen\\
Department of Computer\\Science and Technology,\\Beijing Forestry University\\
	\texttt{fychen98.ai@gmail.com} \\
	\And
	Ziqian Luo\\
	School of Computer Science,\\
	Language Technologies Institute,\\
	Carnegie Mellon University\\
	\texttt{luoziqian98@gmail.com} \\
	\And
	Yanyan Xu\thanks{Corresponding Author}  \\
Department of Computer\\Science and Technology,\\Beijing Forestry University\\
	\texttt{xuyanyan@bjfu.edu.cn} \\
	 \And
	 Dengfeng Ke\thanks{Corresponding Author}  \\
National Laboratory of\\Pattern Recognition,\\Institute of Automation,\\Chinese Academy of Sciences \\
\texttt{dengfeng.ke@nlpr.ia.ac.cn} \\
}

\begin{document}

\maketitle

\begin{abstract}
Sentiment analysis, mostly based on text, has been rapidly developing in the last decade and has attracted widespread attention in both academia and industry. However, information in the real world usually comes from multiple modalities, such as audio and text. Therefore, in this paper, based on audio and text, we consider the task of multimodal sentiment analysis and propose a novel fusion strategy including both multi-feature fusion and multi-modality fusion to improve the accuracy of audio-text sentiment analysis. We call it the DFF-ATMF (Deep Feature Fusion - Audio and Text Modality Fusion) model, which consists of two parallel branches, the audio modality based branch and the text modality based branch. Its core mechanisms are the fusion of multiple feature vectors and multiple modality attention. Experiments on the CMU-MOSI dataset and the recently released CMU-MOSEI dataset, both collected from YouTube for sentiment analysis, show the very competitive results of our DFF-ATMF model. Furthermore, by virtue of attention weight distribution heatmaps, we also demonstrate the deep features learned by using DFF-ATMF are complementary to each other and robust. Surprisingly, DFF-ATMF also achieves new state-of-the-art results on the IEMOCAP dataset, indicating that the proposed fusion strategy also has a good generalization ability for multimodal emotion recognition.
\end{abstract}

% keywords can be removed
%\keywords{Multimodal Fusion\and Multi-Features Fusion \and Sentiment Analysis}

\section{Introduction}

Sentiment analysis provides beneficial information to understand an individual's attitude, behavior, and preference \cite{zhang2018deep}. Understanding and analyzing context-related sentiment is an innate ability of a human being, which is also an important distinction between a machine and a human being \cite{kozinets2018evolving}. Therefore, sentiment analysis becomes a crucial issue in the field of artificial intelligence to be explored.

In recent years, sentiment analysis mainly focuses on textual data, and consequently, text-based sentiment analysis becomes relatively mature \cite{zhang2018deep}. With the popularity of social media such as Facebook and YouTube, many users are more inclined to express their views with audio or video \cite{poria2017review}. Audio reviews become an increasing source of consumer information and are increasingly being followed with interest by companies, researchers and consumers. They also provide more natural experiences than traditional text comments due to allowing viewers to better perceive a commentator's sentiment, belief, and intention through richer channels such as intonation \cite{poria2018combining}. The combination of multiple modalities \cite{zadeh2018multimodal,poria2018combining} brings significant advantages over using only text, including language disambiguation (audio features can help eliminate ambiguous language meanings) and language sparsity (audio features can bring additional emotional information). Also, basic audio patterns can enhance links to the real world environment. Actually, people often associate information with learning and interact with the external environment through multiple modalities such as audio and text \cite{baltruvsaitis2019multimodal}. Consequently, multimodal learning becomes a new effective method for sentiment analysis \cite{majumder2018multimodal}. Its main challenge lies in inferring joint representations that can process and connect information from multiple modalities \cite{poria2018multimodal}.

In this paper, we propose a novel fusion strategy, including the multi-feature fusion and the multi-modality fusion, to improve the accuracy of multimodal sentiment analysis based on audio and text. We call it the DFF-ATMF model, and the learned features have strong complementarity and robustness. We conduct experiments on the CMU Multimodal Opinion-level Sentiment Intensity (CMU-MOSI) \cite{zadeh2016mosi} dataset and the recently released CMU Multimodal Opinion Sentiment and Emotion Intensity (CMU-MOSEI) \cite{zadeh2018multimodal} dataset, both collected from YouTube, and make comparisons with other state-of-the-art models to show the very competitive performance of our proposed model. It is worth mentioning that DFF-ATMF also achieves the most advanced results on the IEMOCAP dataset in the generalized verification experiments, meaning that it has a good generalization ability for multimodal emotion recognition.

The major contributions of this paper are as follows:

\begin{itemize}
	\item We propose the DFF-ATMF model for audio-text sentiment analysis, combining the multi-feature fusion with the multi-modality fusion to learn more comprehensive sentiment information.
	\item The features learned by the DFF-ATMF model have good complementarity and excellent robustness, and even show an amazing performance when generalized to emotion recognition tasks.
	\item Experimental results indicate that the proposed model outperforms the state-of-the-art models on the CMU-MOSI dataset \cite{ghosal2018contextual} and the IEMOCAP dataset \cite{poria2018multimodal}, and also has very competitive results on the recently released CMU-MOSEI dataset.
\end{itemize}

The rest of this paper is structured as follows. In the following section, we review related work. We exhibit the details of our proposed methodologies in Section 3. Then, in Section 4, experimental results and further discussions are presented. Finally, we conclude this paper in Section 5.

\section{Related Work}
	
\subsection{Audio Sentiment Analysis}
Audio data are usually extracted from the characteristics of audio samples' channel, excitation, and prosody. Among them, prosody parameters extracted from segments, sub-segments, and hyper-segments are used for sentiment analysis in \cite{liu2018speech}. In the past several years, classical machine learning algorithms, such as Hidden Markov Model (HMM), Support Vector Machine (SVM), and decision tree-based methods, have been utilized for audio sentiment analysis \cite{schuller2004speech,schuller2003hidden,lee2011emotion}. Recently, researchers have proposed various neural network-based architectures to improve audio sentiment analysis. In 2014, an initial study employed deep neural networks (DNNs) to extract high-level features from raw audio data and demonstrated its effectiveness \cite{han2014speech}. With the development of deep learning, more complex neural-based architectures have been proposed. For example, convolutional neural network (CNN)-based models have been used to train spectrograms or audio features derived from original audio signals such as Mel Frequency Cepstral Coefficients (MFCCs) and Low-Level Descriptors (LLDs) \cite{bertero2017first,parthasarathy2018convolutional,minaee2019deep}.

\subsection{Text Sentiment Analysis}
After decades of development, text sentiment analysis has become mature in recent years \cite{hussein2018survey}. The most commonly used classification techniques such as SVM, maximum entropy and naive Bayes, are based on the word bag model, where the sequence of words is ignored, which may result in inefficient extraction of sentiment from the input because the sequence of words will affect the existing sentiment \cite{chaturvedi2018distinguishing}. Later research has overcome this problem by using deep learning in sentiment analysis \cite{zhang2018deep}. For instance, a kind of DNN model is proposed, using word-level, character-level and sentence-level representations for sentiment analysis \cite{jianqiang2018deep}. In order to better capture the temporal information, \cite{dai2019transformer} proposes a novel neural architecture, called Transformer-XL, which enables learning dependency beyond a fixed-length without disrupting temporal coherence. It consists of a segment-level recurrence mechanism and a novel positional encoding scheme, not only capturing longer-term dependency but also resolving the context fragmentation problem.

\subsection{Multimodal Learning}
Multimodal learning is an emerging field of research \cite{baltruvsaitis2019multimodal}. Learning from multiple modalities needs to capture the correlation among these modalities. Data from different modalities may have different predictive power and noise topology, with possibly losing the information of at least one of the modalities \cite{baltruvsaitis2019multimodal}. \cite{majumder2018multimodal} presents a novel feature fusion strategy that proceeds in a hierarchical manner for multimodal sentiment analysis. \cite{ghosal2018contextual} proposes a recurrent neural network-based multimodal attention framework that leverages contextual information for utterance-level sentiment prediction and shows a state-of-the-art model on the CMU-MOSI and CMU-MOSEI datasets.

\begin{figure}[htbp] 
	% 		\centering
	\includegraphics[width=0.45\textwidth]{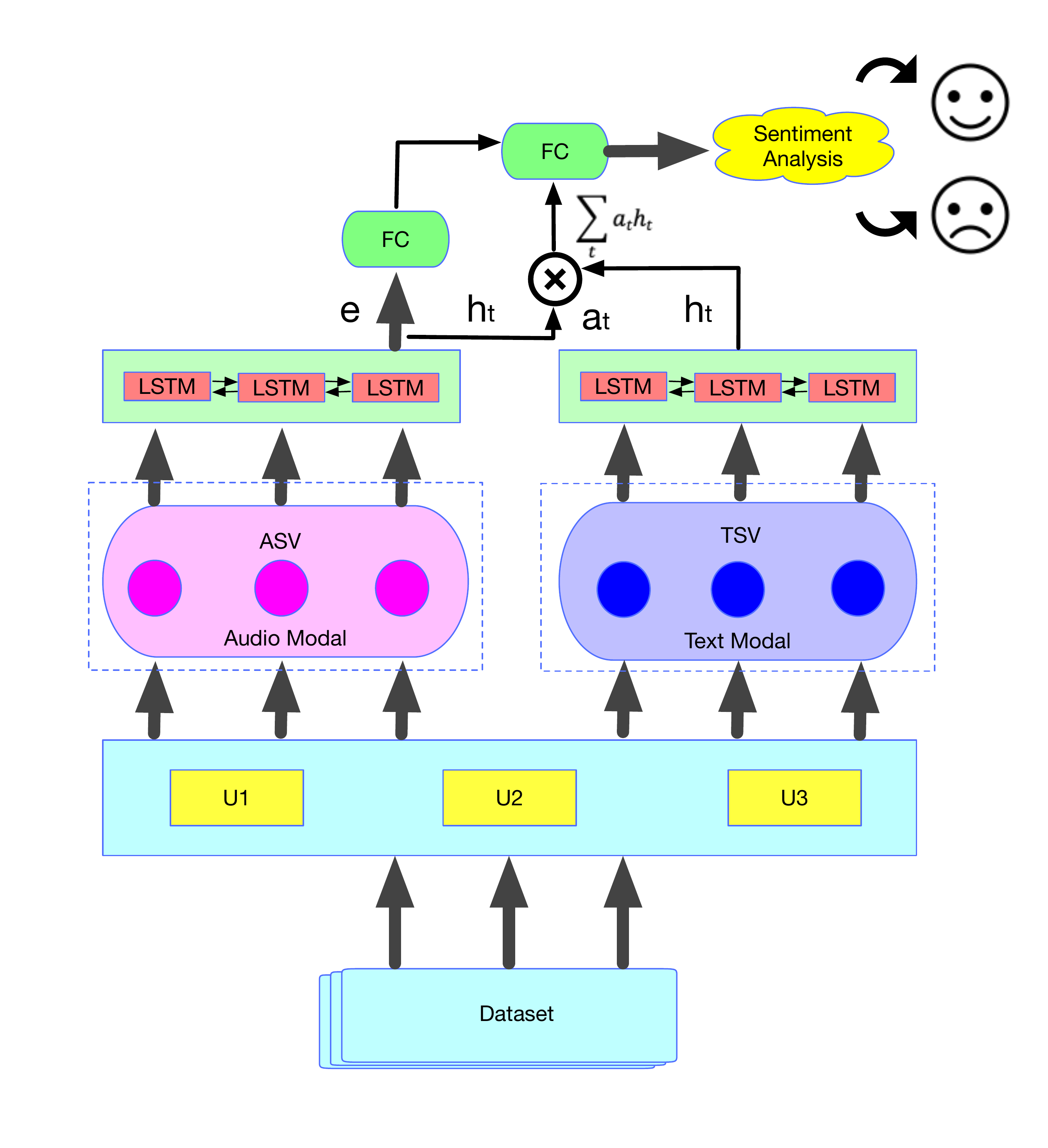} 
	\caption{The overall architecture of the proposed DFF-ATMF framework. $h_t$ represents the hidden state of Bi-LSTM at time $t$. $e$ means the final audio sentiment vector. $a_t$ represents the attention weight and is calculated as the dot product of the final audio sentiment vector and the final text sentiment vector of $h_t$.  ``FC'' means a fully-connected layer. }
	\label{Fig.main1} 
\end{figure}

\section{Proposed Methodology}
	
In this section, we describe the proposed DFF-ATMF model for audio-text sentiment analysis in detail. We firstly introduce an overview of the whole neural network architecture, illustrating how to fuse audio and text modalities. After that, two separate branches of DFF-ATMF are respectively explained to show how to fuse the audio feature vector and the text feature vector. Finally, we present the multimodal-attention mechanism used in the DFF-ATMF model.

\subsection{The DFF-ATMF Framework}
The overall architecture of the proposed DFF-ATMF framework is shown in Figure \ref{Fig.main1}. We fuse audio and text modalities in this framework through two parallel branches, that is, the audio modality based branch and the text modality based branch. DFF-ATMF's core mechanisms are feature vector fusion and multimodal-attention fusion. The audio modality branch uses Bi-LSTM \cite{cai2018multi} to extract audio sentiment information between adjacent utterances (U1, U2, U3), while another branch uses the same network architecture to extract text features. Furthermore, the audio feature vector of each piece of utterance is used as the input of our proposed neural network, which is based on the audio feature fusion, so we can obtain a new feature vector before the softmax layer, called the audio sentiment vector (ASV). The text sentiment vector (TSV) can be achieved similarly. Finally, after the multimodal-attention fusion, the output of the softmax layer produces final sentiment analysis results, as shown in Figure \ref{Fig.main1}.

\begin{figure}[htbp] 
%	\centering
	\includegraphics[width=0.45\textwidth]{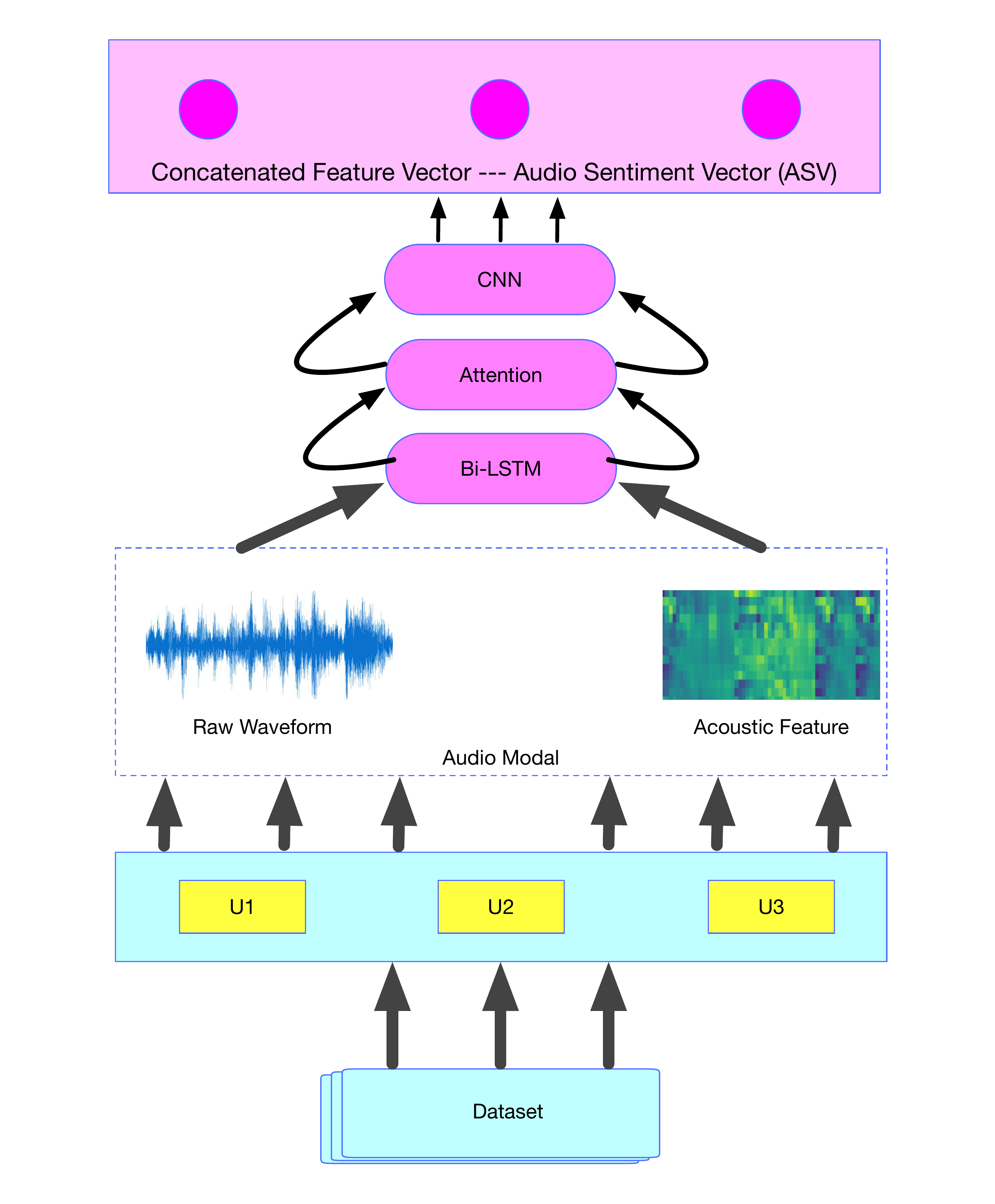} 
	\caption{The architecture of ASV from AFF.} 
	\label{Fig.main2} 
\end{figure}

\subsection{Audio Sentiment Vector (ASV) from Audio Feature Fusion (AFF)}

Base on the work in \cite{luo2019audio}, in order to explore further the fusion of feature vectors inter the audio modality, we extend the experiments of different types of audio features on the CMU-MOSI dataset, and the results are shown in Table \ref{Tab.main1}.

	\begin{table*}[ht]
		\setlength{\belowcaptionskip}{10pt}
		
		\centering
		\fontsize{9}{10}\selectfont
		\setlength{\abovecaptionskip}{15pt}
		\setlength{\tabcolsep}{6.4mm}
		\caption{Comparison of different types of audio features on the CMU-MOSI dataset.}
		\begin{tabular}{lllll}
			\hline
			\multirow{2}{*}{Feature}                   & \multirow{2}{*}{Model} & \multicolumn{3}{c}{Accuracy(\%)}                 \\ \cline{3-5} 
			&                        & 2-class        & 5-class        & 7-class        \\ \hline
			\multirow{2}{*}{1 Chromagram from spectrogram (chroma\_stft)}          & LSTM                   & 43.24          & 20.23          & 13.96          \\
			& BiLSTM                 & 45.37          & 2.29           & 12.39          \\
			\multirow{2}{*}{2 Chroma Energy Normalized (chroma\_cens)}             & LSTM                   & 42.98          & 20.87          & 13.31          \\
			& BiLSTM                 & 45.85          & 20.53          & 13.76          \\
			\multirow{2}{*}{\textbf{3 Mel-frequency cepstral coefficients (MFCC)}} & \textbf{LSTM}          & \textbf{55.12} & \textbf{23.64} & \textbf{16.99} \\
			& \textbf{BiLSTM}        & \textbf{55.98} & \textbf{23.75} & \textbf{17.24} \\
			\multirow{2}{*}{4 Root-Mean-Square Energy (RMSE)}                      & LSTM                   & 52.30          & 21.14          & 15.33          \\
			& BiLSTM                 & 52.76          & 22.35          & 15.87          \\
			\multirow{2}{*}{5 Spectral\_Centroid}                                  & LSTM                   & 48.39          & 22.25          & 14.97          \\
			& BiLSTM                 & 48.84          & 22.36          & 15.79          \\
			\multirow{2}{*}{6 Spectral\_Contrast}                                  & LSTM                   & 48.34          & 22.50          & 15.02          \\
			& BiLSTM                 & 48.97          & 22.28          & 15.98          \\
			\multirow{2}{*}{7 Tonal Centroid Features (tonnetz)}                   & LSTM                   & 53.78          & 22.67          & 15.83          \\
			& BiLSTM                 & 54.24          & 21.87          & 16.01          \\ \hline
		\end{tabular}
		\label{Tab.main1}
	\end{table*}

In addition, we also implement an improved serial neural network of Bi-LSTM and CNN \cite{wu2018thu_ngn}, combining with the attention mechanism to learn the deep features of different sound representations. The multi-feature fusion procedure is described with the LSTM branch and the CNN branch respectively in Algorithm \ref{Alg.main1}. 

	\begin{algorithm}[ht]  
	\caption{The Multi-Feature Fusion Procedure}  
	\begin{algorithmic}[1]  
		\Procedure{LSTM branch}{}
		\For{i:[0,n]}  
		\State $f_i=getAudioFeature(u_i)$  // get the audio feature from the $u^{th}$ utterance
		\State $a_i=getASV(f_i)$   
		\EndFor 
		\For{i:[0,M]}               //M is the number of videos
		\State $input_i=GetTopUtter(v_i)$  
		\State $u_{f_i}=getUtterFeature(input_i)$   
		\EndFor  
		\State $shuffle(v)$
		\State $Attention(A_i)$
		\State $Multi-Feature$ $Fusion$ $from$ $the$ $LSTM$ $branch$
		\EndProcedure
		\Procedure{CNN Branch}{}
		\For{i:[0,n]}  
		\State $\textit{$x_i$} \gets \textit{get SpectrogramImage($u_i$)}$ 
		\State $\textit{$c_i$} \gets 	\textit{CNNModel($x_i$)}$
		\EndFor  
		\State $Attention(C_i)$
		\State $Multi-Feature$ $Fusion$ $from$ $the$ $CNN$ $branch$
		\EndProcedure
		
		\Procedure{Feature Fusion}{}
		\For{i:[0,n]}  
		\State $L_i=Attention(a_i)$
		\State $C_i=Attention(l_i)$
		\EndFor  
		\State $Attention(L_i + C_i)$
		\State $Multi-Feature$ $Fusion$
		\EndProcedure
	\end{algorithmic}  
	\label{Alg.main1}
\end{algorithm}

The features are learned from raw waveforms and acoustic features, which are complementary to each other. Therefore, audio sentiment analysis can be improved by applying our feature fusion technique, that is, ASV from AFF, whose architecture is shown in Figure \ref{Fig.main2}.

\begin{figure}[htbp] 
	\centering
	\includegraphics[width=0.46\textwidth]{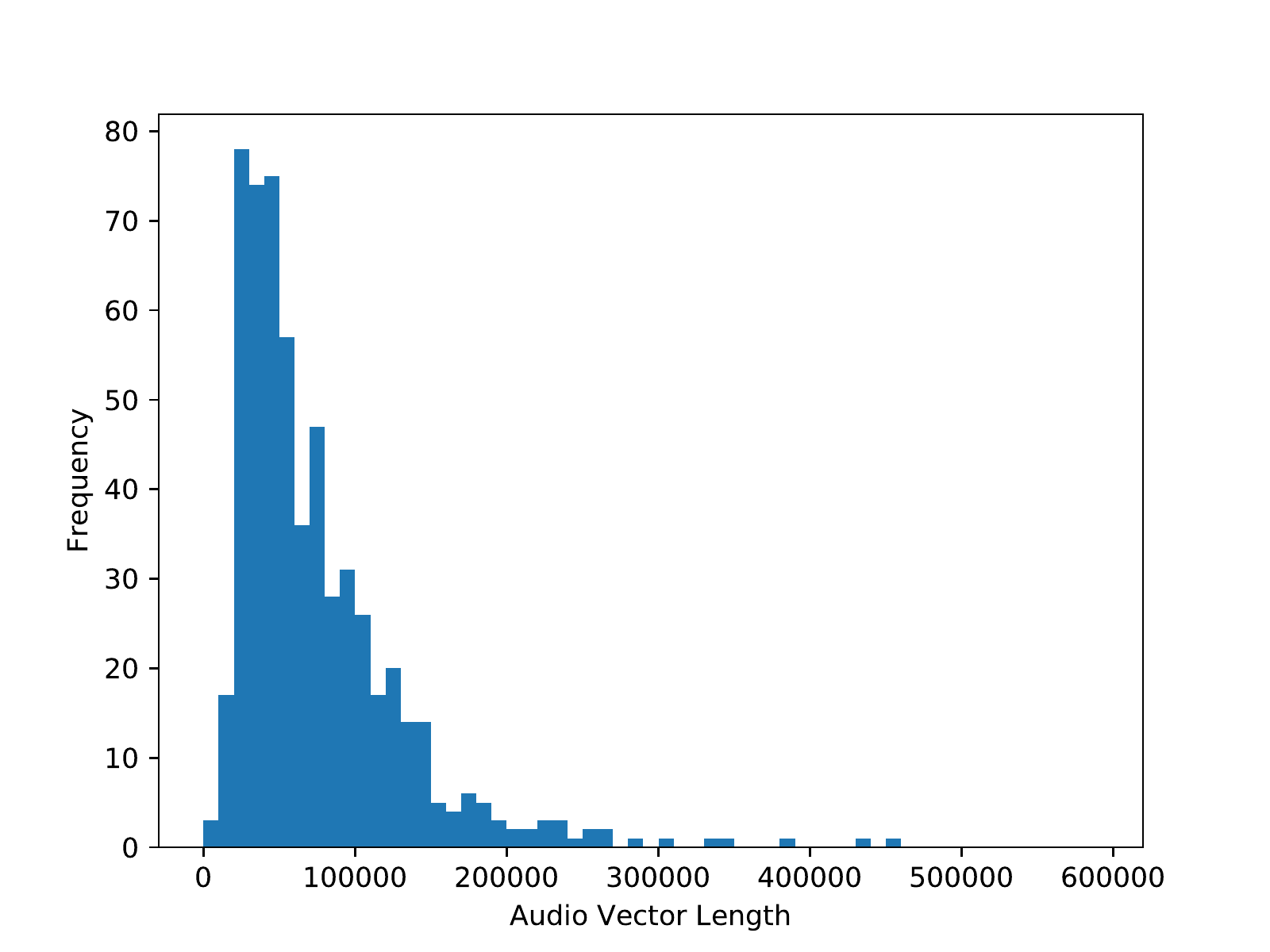}
	\caption{The raw audio waveform sampling distribution on the CMU-MOSI dataset.} 
	\label{Fig.main3} 
\end{figure}

In terms of raw audio waveforms, taking the CMU-MOSI dataset as an example, we illustrate their sampling distribution in Figure \ref{Fig.main3}. The inputs to the network are raw audio waveforms sampled at 22 kHz. We also scale the waveforms to be in the range [-256, 256], so that we do not need to subtract the mean value as the data are naturally near zero already. To obtain a better sentiment analysis accuracy, batch normalization (BN) and the ReLU function are employed after each convolutional layer. Additionally, dropout regularization is also applied to the proposed serial network architecture.

In terms of acoustic features, we extract them using the Librosa \cite{mcfee2015librosa} toolkit and obtain four effective kinds of features to represent sentiment information, which are MFCCs, spectral\_centroid, chroma\_stft and spectral\_contrast, respectively. In particular, taking log-Mel spectrogram extraction \cite{yin2018learning} as an example, we use 44.1 kHz without downsampling and extract the spectrograms with 64 Bin Mel-scale. The window size for short-time Fourier transform is 1,024 with a hop size of 512. The resulting Mel-spectrograms are next converted into log-scaled ones and standardized by subtracting the mean value and divided by the standard deviation.

Finally, we feed feature vectors of raw waveforms and acoustic features into our improved serial neural network of Bi-LSTM and CNN, combining with the attention mechanism to learn the deep features of different sound representations, that is, ASV.

\begin{figure}[htbp] 
	\centering
	\includegraphics[width=0.45\textwidth]{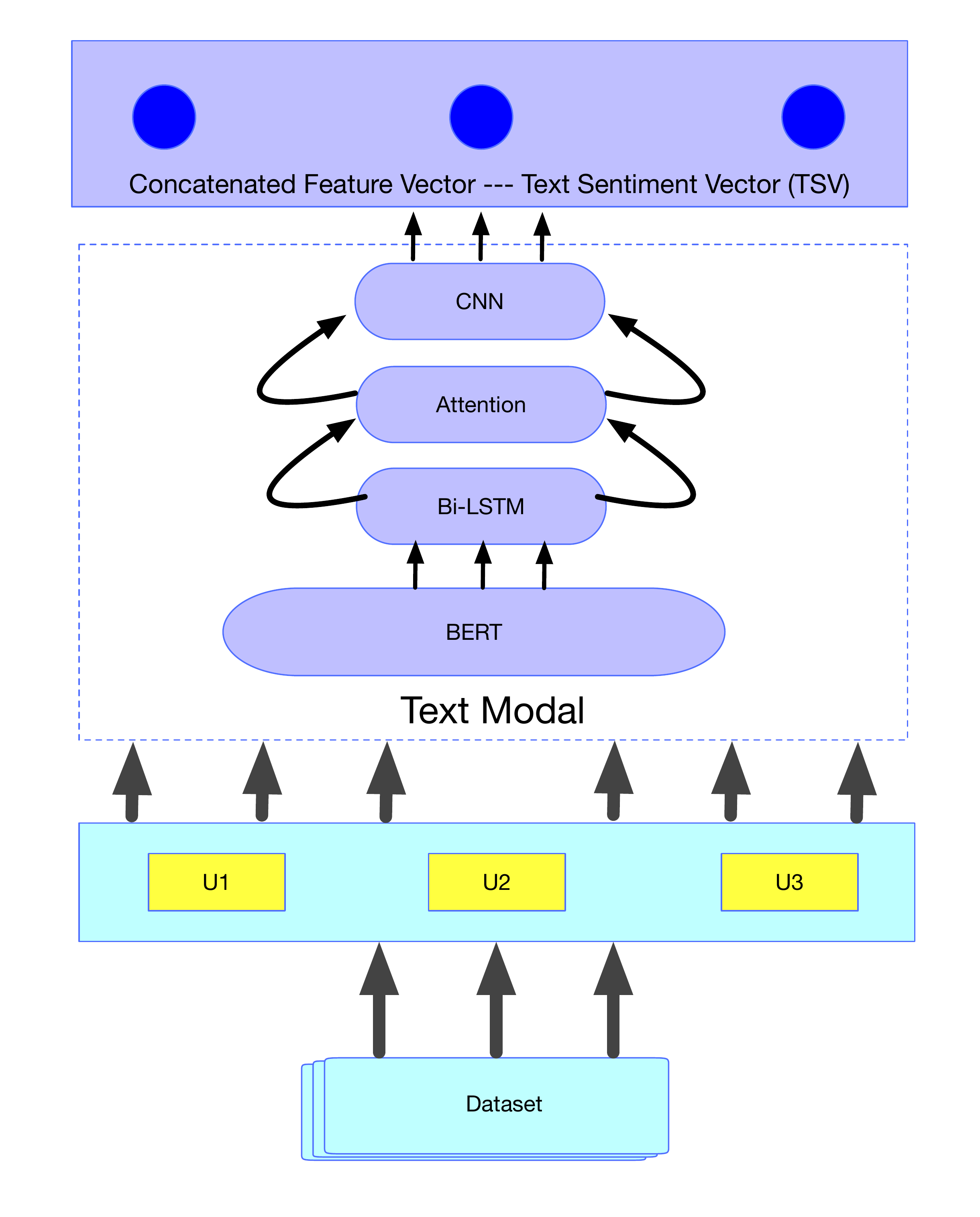} 
	\caption{The architecture of TSV from TFF.} 
	\label{Fig.main4} 
\end{figure}

\subsection{Text Sentiment Vector (TSV) from Text Feature Fusion (TFF)}

	The architecture of TSV from TFF is shown in Figure \ref{Fig.main4}. BERT \cite{devlin2018bert} is a new language representation model, standing for Bidirectional Encoder Representations from Transformers. Thus far, to the best of our knowledge, no studies have leveraged BERT to pre-train text feature representations on the multimodal dataset such as CMU-MOSI. We then utilize BERT embeddings for CMU-MOSI. Next, the Bi-LSTM layer takes the concatenated word embeddings and POS tags as its inputs and it outputs each hidden state. Let $h_i$ be the output hidden state at time $i$. Then its attention weight $a_i$ can be formulated by Equation \ref{eqn1}.
	
\begin{equation} 
\begin{split}
m_i=tanh(h_i)     \\
\hat{a _i}=w_im_i+b_i    \\
a _i = \frac{exp(\hat{a _i})}{\sum _jexp(\hat{a _j})}
\end{split}
\label{eqn1}
\end{equation}

In Equation \ref{eqn1}, $w_im_i + b_i$ denotes a linear transformation of $m_i$. Therefore, the output representation $r_i$ is given by:

\begin{equation} 
r_i = a _ih_i.
\label{eqn2}
\end{equation}

	Based on such text representations, the sequence of features will be assigned with different attention weights. Thus, crucial information such as emotional words can be identified more easily. The convolutional layer takes the text representation $r_i$ as its input, and the output CNN feature maps are concatenated together. Finally, text sentiment analysis can be improved by using TSV from TFF.
	
	\subsection{Audio and Text Modal Fusion with the Multimodal-Attention Mechanism} 
	
	Inspired by human visual attention, the attention mechanism, proposed by \cite{luong2015effective} for neural machine translation, is introduced into the encoder-decoder framework to select reference words from the source language for the words in the target language. Based on the existing attention mechanism, inspired by the work in \cite{yoon2018multimodal}, we improve the multimodal-attention method on the basis of the multi-feature fusion strategy, focusing on the fusion of comprehensive and complementary sentiment information from audio and text. We leverage the multimodal-attention mechanism to preserve the intermediate outputs of the input sequences by retaining the Bi-LSTM encoder, and then a model is trained to selectively learn these inputs and to correlate output sequences with the model's output.
	
	More specifically, ASV and TSV are firstly encoded with Audio-BiLSTM and Text-BiLSTM using Equation \ref{eqn3}.
	
	\begin{equation}
	\begin{split}
	A_{t+1} = f_\theta (A_{t},x_{t+1}) \\
	A_{t-1} = f_\theta (A_{t},x_{t-1}) \\
	T_{t+1} = f_\theta (T_{t},x_{t+1}) \\
	T_{t-1} = f_\theta (T_{t},x_{t-1})
	\end{split}
	\label{eqn3}
	\end{equation}
	
	In Equation \ref{eqn3}, $f_\theta$ is the LSTM function with the weight parameter $\theta$. $A_{t+1}$, $A_t$ and $A_{t-1}$ represent the hidden states at time ${(t+1)}^{th}$, ${t}^{th}$ and ${(t-1)}^{th}$ from the audio modality, respectively. $x_{t+1}$ and $x_{t-1}$ represent the features at time ${(t+1)}^{th}$ and ${(t-1)}^{th}$, respectively. The text modality is similar, represented by $T$.
	
	\begin{equation}
	\begin{split}
	a _t= \frac{exp(e^Th_t)}{\sum _texp(e^Th_t)}\\
	t _t= \frac{exp(e^Th'_t)}{\sum _texp(e^Th'_t)}\\   
	Z_a = \sum _ta_th_t\\ 
	Z_t = \sum _tt_th'_t\\
	\end{split}
	\label{eqn4}
	\end{equation}
	
	\begin{equation} 
	\begin{split}
	\hat{y}_{i,j} = softmax(concat(concat(Z_a,Z_t),A)^TM+b)
	\end{split}
	\label{eqn5}
	\end{equation}
	
	We then consider the final ASV $e$ as an intermediate vector, as shown in Figure \ref{Fig.main1}. During each time step $t$, the dot product of the intermediate vector $e$ and the hidden state $h_t$ is evaluated to calculate a similarity score $a_t$. Using this score as a weight parameter, the weighted sum $\sum _ta_th_t$ is calculated to generate a multi-feature fusion vector $Z_a$. The multi-feature fusion vector of the text modality is calculated similarly, represented by $Z_t$. We are therefore able to obtain two kinds of multi-feature fusion vectors for the audio modality and the text modality respectively, as shown in Equation \ref{eqn4} and \ref{eqn5}. These multi-feature fusion vectors are respectively concatenated with the final intermediate vectors of ASV and TSV, which will pass through the softmax function to perform sentiment analysis, as shown in Equation \ref{eqn6} and \ref{eqn7}.
	
	\begin{equation} 
	\begin{split}
ASV = g_\theta (e)  \\   
TSV = g{_\theta }'(h_t)     \\
	\end{split}
	\label{eqn6}
	\end{equation}
	
	\begin{equation} 
	\begin{split}
	\hat{y}_{i} = softmax(concat(ASV,TSV)^TM+b)
	\end{split}
	\label{eqn7}
	\end{equation}

	\begin{table*}[tp]
		
		\centering
		\fontsize{9}{10}\selectfont
		\setlength{\abovecaptionskip}{5pt}
		\setlength{\tabcolsep}{6.4mm}
		\caption{Datasets for training and test in our experiments.}
		{
			\begin{threeparttable}
				
				\label{tab:performance_comparison}
				\begin{tabular}{p{2cm} p{1cm}<{\centering} p{1cm}<{\centering} p{1cm}<{\centering} p{1cm}<{\centering}}
					\toprule
					\multirow{2}{*}{Dataset}&
					\multicolumn{2}{c}{ Training}&\multicolumn{2}{c}{ Test}\cr
					\cline{2-5}
					&\#utterance&\#video&\#utterance&\#video\cr
					\hline
					CMU-MOSI&1 616&65&583&28\cr
					CMU-MOSEI&18 051&1 550&4 625&679\cr
					IEMOCAP&4 290&120&1 208&31\cr
					\bottomrule
				\end{tabular}
		\end{threeparttable}}
		
	\end{table*}
	\label{Tab.main2}
	\section{Empirical Evaluation}
	
	In this section, we firstly introduce the datasets, the evaluation metrics and the network structure parameters used in our experiments, and then exhibit the experimental results and make comparisons with other state-of-the-art models to show the advantages of DFF-ATMF. At last, more discussions are illustrated to understand the learning behavior of DFF-ATMF better.
		 
	\subsection{Experiment Settings}
	
	\subsubsection{Datasets}

The datasets used for training and test are depicted in Table \ref{tab:performance_comparison}. The CMU-MOSI dataset is rich in sentiment expression, consisting of 2,199 utterances, that is, 93 videos by 89 speakers. The videos involve a large array of topics such as movies, books, and other products. These videos were crawled from YouTube and segmented into utterances where each utterance is annotated with scores between $-3$ (strongly negative) and +3 (strongly positive) by five annotators. We take the average of these five annotations as the sentiment polarity and then consider only two classes, that is, ``positive'' and ``negative''. Our training and test splits of the dataset are completely disjoint with respect to speakers. In order to better compare with the previous work, similar to \cite{poria2018multimodal}, we divide the dataset by 7:3 approximately, resulting in 1,616 and 583 utterances for training and test respectively. 

The CMU-MOSEI dataset is an upgraded version of the CMU-MOSI dataset, which has 3,229 videos, that is, 22,676 utterances, from more than 1,000 online YouTube speakers. The training and test sets include 18,051 and 4,625 utterances respectively, similar to \cite{ghosal2018contextual}. 

The IEMOCAP dataset was collected following theatrical theory in order to simulate natural dyadic interactions between actors. We use categorical evaluations with majority agreement and use only four emotional categories, that is, ``happy'', ``sad'', ``angry'', and ``neutral'' to compare the performance of our model with other researches using the same categories \cite{poria2018multimodal}. 
	
	\subsubsection{Evaluation Metrics} 
	
	We evaluate the performance of our proposed model by the \textit{weighted accuracy} on 2-class or multi-class classifications. 
	
	\begin{equation}
	weighted{\kern 1pt} {\kern 1pt} {\kern 1pt} {\kern 1pt} accuracy = \frac{{correct{\kern 1pt} {\kern 1pt} {\kern 1pt} {\kern 1pt} utterances}}{{utterances}}  
	\label{eqn8}
	\end{equation}

	Additionally, F1-Score is used to evaluate 2-class classification.
	
	\begin{equation}
	{{\rm{F}}_\beta }{\rm{ = }}(1 + {\beta ^2}) \cdot \frac{{precision \cdot recall}}{{({\beta ^2} \cdot precision) + recall}}
	\label{eqn9}
	\end{equation}
	
	In Equation \ref{eqn9}, $\beta$ represents the weight between precision and recall. During our evaluation process, we set $\beta$ = 1 since we consider precision and recall to have the same weight, and thus $F1$-score is adopted.
	
	However, in emotion recognition, we use Macro $F1$-Score to evaluate the performance.
	
	\begin{equation}
	Macro{\kern 1pt} {\kern 1pt} {\kern 1pt} {\kern 1pt} {F1}{\rm{ = }}\frac{{\sum\limits_1^n {{F_{1n}}} }}{n}
	\label{eqn10}
	\end{equation}
	
	In Equation \ref{eqn10}, $n$ represents the number of classifications and $F_{1n}$ is the $F1$ score on the $n^{th}$ category.
		
	\subsubsection{Network Structure Parameters}
	
	Our proposed architecture is implemented on the open-source deep learning framework TensorFlow. More specifically, for the proposed audio and text multi-modality fusion framework, we use Bi-LSTM with $200$ neurons, each followed by a dense layer consisting of $100$ neurons. Utilizing the dense layer, we project the input features of audio and text to the same dimension, and next combine them with the multimodal-attention mechanism. We set the dropout hyperparameter to be $0.4$ for CMU-MOSI and $0.3$ for CMU-MOSEI \& IEMOCAP as a measure of regularization. We also use the same dropout rates for the Bi-LSTM layers. We employ the ReLu function in the dense layers and softmax in the final classification layer. When training the network, we set the batch size to be $32$, and use Adam optimizer with the cross-entropy loss function and train for $50$ epochs. In data processing, we make each utterance one-to-one correspondence with the label and rename the utterance.
	
	The network structure of the proposed audio and text multi-feature fusion framework is similar. Taking the audio multi-feature fusion framework as an example, the hidden states of Bi-LSTM are of $2*200$-dim. The kernel sizes of CNN are $3$, $5$, $7$ and $9$ respectively. The size of the feature map is $4*200$. A dropout rate is a random number between $0.3$ and $0.4$. The loss function used is MAE, and the batch size is set to $16$. We combine the training set and the development set in our experiments. We use 90\% for training and reserve 10\% for cross-validation. To train the feature encoder, we follow the fine-tuning training strategy. 
	
	In order to reduce randomness and improve credibility, we report the average value over $3$ runs for all experiments.
	
	\subsection{Experimental Results}
	
	\subsubsection{Comparison with Other Models}
		
		\begin{table*}[ht]
		\setlength{\belowcaptionskip}{10pt}
		
		\centering
		\fontsize{9}{10}\selectfont
		\setlength{\abovecaptionskip}{15pt}
		\setlength{\tabcolsep}{4mm}
		\caption{Comparison with other state-of-the-art models.}
		{
			\begin{tabular}{c|l|l|c|c|c|c|cc}
				\hline
				\multicolumn{3}{c|}{\multirow{2}{*}{Model}} & \multicolumn{2}{c|}{CMU-MOSI} & \multicolumn{2}{c|}{CMU-MOSEI} & \multicolumn{2}{c}{IEMOCAP}             \\ \cline{4-9} 
				\multicolumn{3}{c|}{}                         & Acc(\%)        & F1          & Acc(\%)      & F1      & \multicolumn{1}{c|}{Overall Acc(\%)} & Macro F1 \\ \hline
				\multicolumn{3}{c|}{\cite{poria2017context}}                    & 79.30          & 80.12       & -        & -         & \multicolumn{1}{c|}{75.60}   & 76.31    \\
				\multicolumn{3}{c|}{\cite{poria2017multi}}                    & 80.10          & 80.62       & -        & -         & \multicolumn{1}{c|}{-}   & -    \\
				\multicolumn{3}{c|}{\cite{zadeh2018multimodal}}                    & 74.93          & 75.42       & 76.24        & 77.03         & \multicolumn{1}{c|}{-}   & -    \\
				\multicolumn{3}{c|}{\cite{poria2018multimodal}}                 & 76.60          & 76.93       & -        & -         & \multicolumn{1}{c|}{78.20}   & 78.79    \\
				\multicolumn{3}{c|}{\cite{ghosal2018contextual}}                 & 80.58          & 80.96       & 79.74        & 80.15        & \multicolumn{1}{c|}{-}   & -    \\
				\multicolumn{3}{c|}{\cite{lee2018convolutional}}                  & -          & -       & \textbf{84.08}        & \textbf{88.89}         & \multicolumn{1}{c|}{-}   & -    \\
				\multicolumn{3}{c|}{\textbf{DFF-ATMF}}                & \textbf{80.98}          & \textbf{81.26}       & 77.15       & 78.33         & \multicolumn{1}{c|}{\textbf{81.37}}   & \textbf{82.29}   \\ \hline
		\end{tabular}}
		\label{Tab.main3}
	\end{table*}
	
	\begin{table*}[ht]
		\centering
		\fontsize{9}{10}\selectfont
		\setlength{\abovecaptionskip}{15pt}
		\setlength{\tabcolsep}{18mm}
		\caption{Experimental results on the IEMOCAP dataset.}
		{
			\begin{tabular}{c|c|c}
				\hline
				\multirow{2}{*}{Emotion} & \multicolumn{2}{c}{IEMOCAP}       \\ \cline{2-3} 
				& ACC(\%)        & Macro F1             \\ \hline
				happy                   & 74.41          & 75.66          \\
				sad            & 73.62          & 74.31          
				\\
				angry                    & 78.57          & 79.14          \\
				neutral                    & 64.35          & 65.72          \\
				\textbf{Overall}                    & \textbf{81.37}          & \textbf{82.29}          
				\\\hline
		\end{tabular}}
		\label{Tab.main4}
	\end{table*}
	
		\begin{itemize}
		\item \cite{poria2017context} proposes an LSTM-based model that enables utterances to capture contextual information from their surroundings in the video, thus aiding the classification. 
\item \cite{poria2017multi} introduces attention-based networks to improve both context learning and dynamic feature fusion.
\item \cite{zadeh2018multimodal} proposes a novel multimodal fusion technique called Dynamic Fusion Graph (DFG).
\item \cite{poria2018multimodal} explores three different deep learning-based architectures, each improving upon the previous one, \textbf{which is the state-of-the-art method on the IEMOCAP dataset at present.}
\item \cite{ghosal2018contextual} proposes a recurrent neural network-based multimodal-attention framework that leverages the contextual information, \textbf{which is the state-of-the-art model on the CMU-MOSI dataset at present.}
\item \cite{lee2018convolutional} proposes a new method of learning about the hidden representations between speech and text data using CNN, \textbf{which is the state-of-the-art model on the CMU-MOSEI dataset at present.}
	\end{itemize}
	
	Table \ref{Tab.main3} shows the comparison of DFF-ATMF with other state-of-the-art models. From Table \ref{Tab.main3}, we can see that DFF-ATMF outperforms the other models on the CMU-MOSI dataset and the IEMOCAP dataset. At the same time, the experimental results on the CMU-MOSEI dataset also show DFF-ATMF's competitive performance.

		\subsubsection{Generalization Ability Analysis}
	
	In order to verify the feature complementarity of our proposed fusion strategy and its robustness, we conduct experiments on the IEMOCAP dataset to examine DFF-ATMF's generalization capability. Surprisingly, our proposed fusion strategy is effective on the IEMOCAP dataset and outperforms the current state-of-the-art method in \cite{poria2018multimodal}, which can be seen from Table \ref{Tab.main3} and the overall accuracy is improved by 3.17\%. More detailed experimental results on the IEMOCAP dataset are illustrated in Table \ref{Tab.main4}.
	
	\subsection{Further Discussions}

	\begin{figure}[htbp] 
		\centering
		\includegraphics[width=0.4\textwidth]{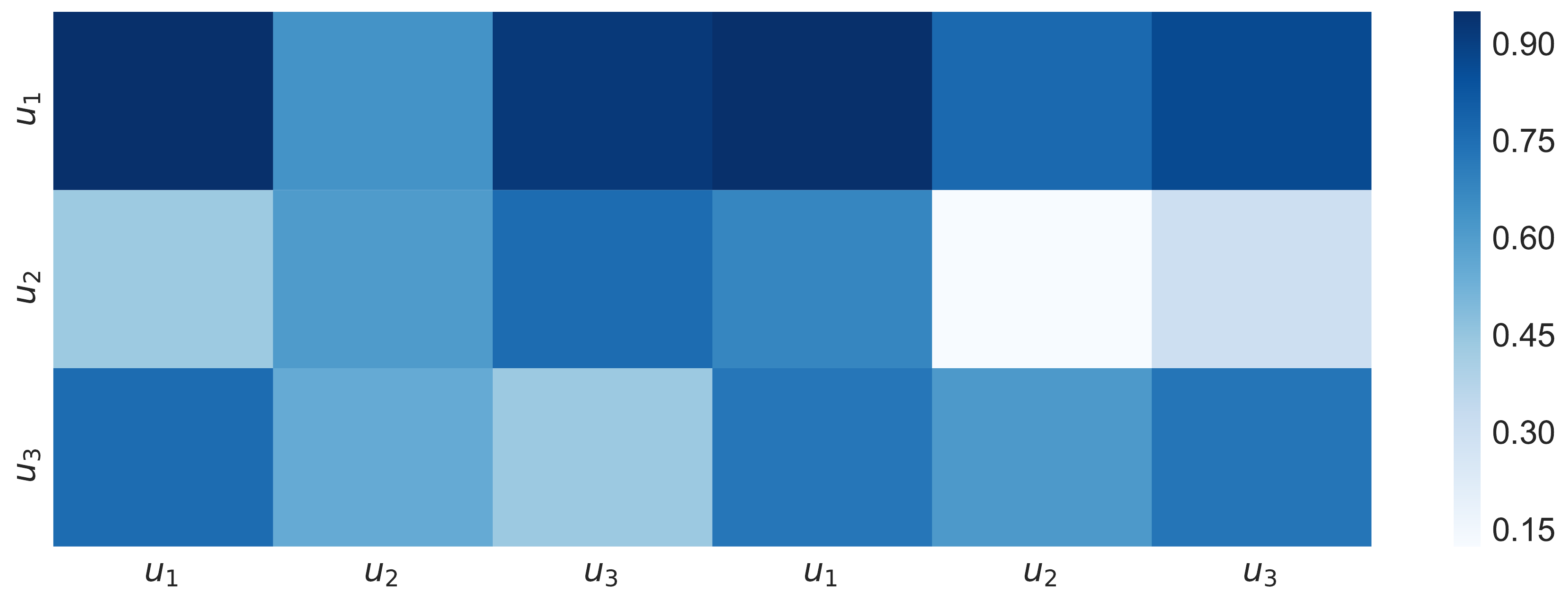}
		\caption{Softmax attention weights of an example from the CMU-MOSI test set.} 
		\label{Fig.main5} 
	\end{figure}

    \begin{figure}[htbp] 
		\centering
		\includegraphics[width=0.4\textwidth]{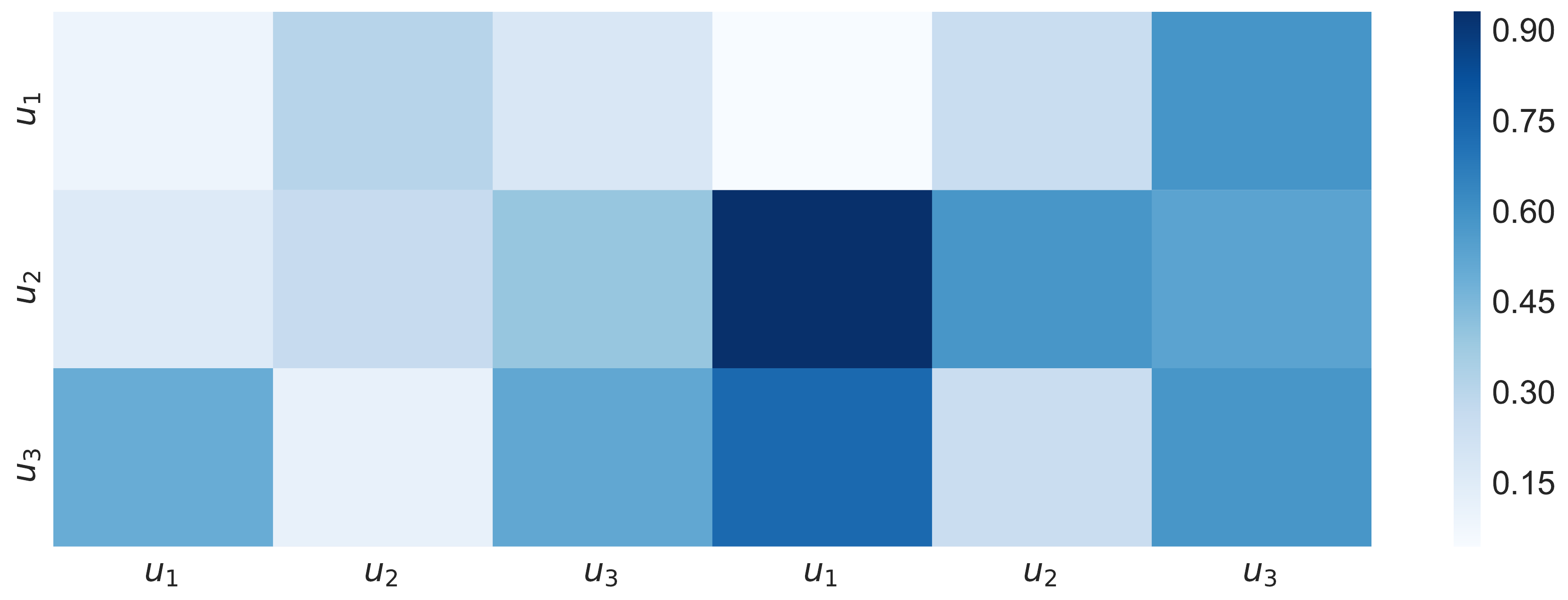}
		\caption{Softmax attention weights of an example from the CMU-MOSEI test set.} 
		\label{Fig.main6} 
	\end{figure}
	
	\begin{figure}[htbp] 
		\centering
		\includegraphics[width=0.4\textwidth]{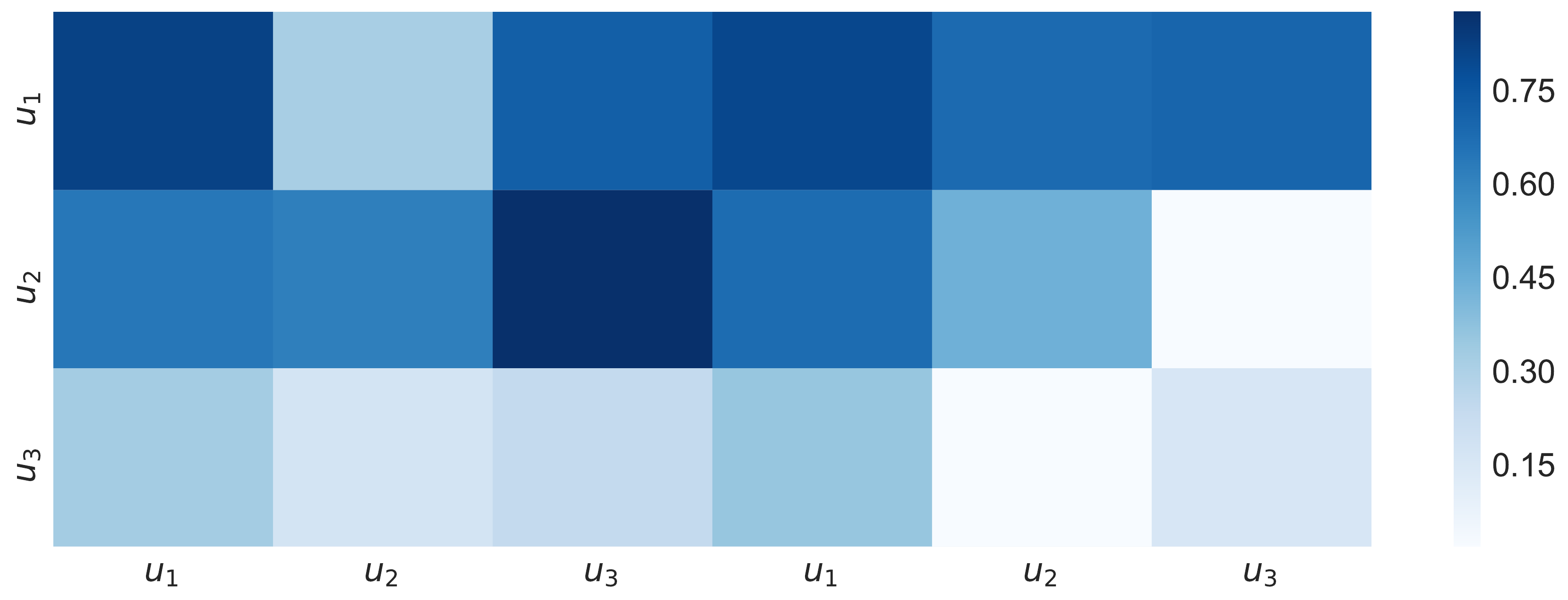}
		\caption{Softmax attention weights of an example from the IEMOCAP test set.} 
		\label{Fig.main7} 
	\end{figure}
		
	\begin{figure}[htbp] 
		\centering
		\includegraphics[width=0.4\textwidth]{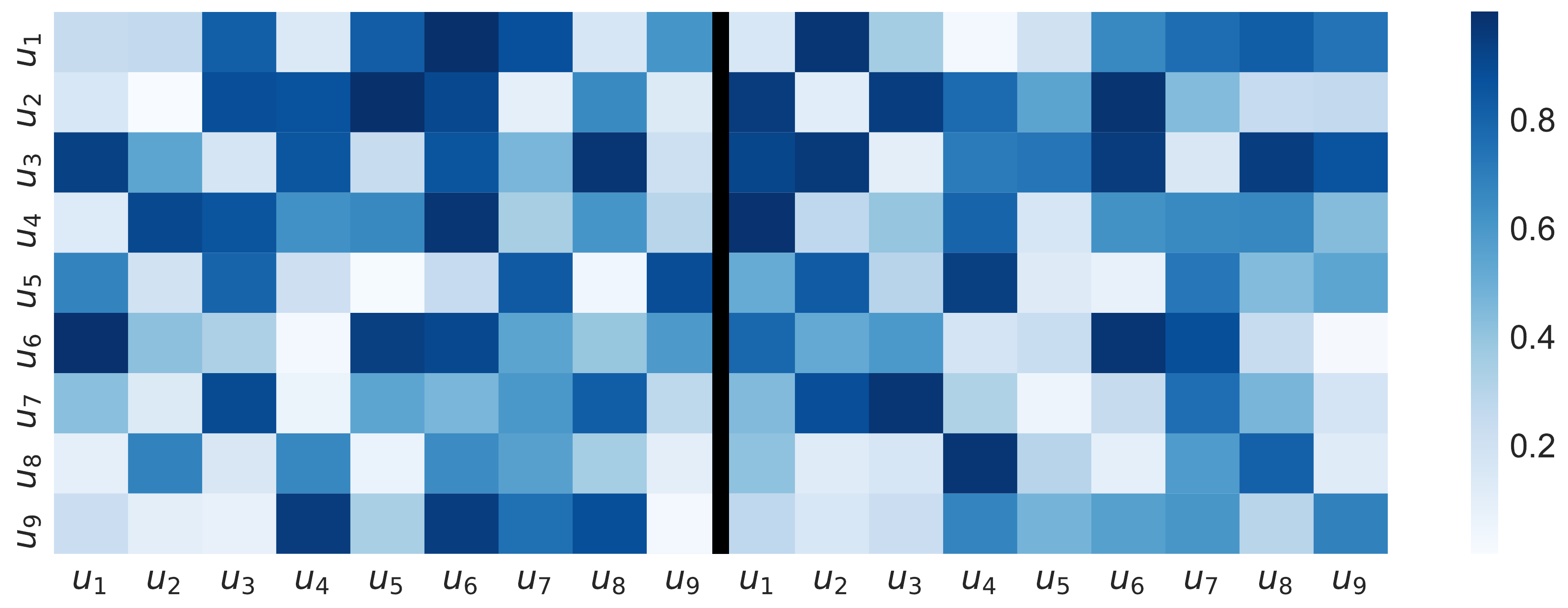}
		\caption{Softmax attention weight comparison of the CMU-MOSI, CMU-MOSEI, and IEMOCAP test sets.} 
		\label{Fig.main8} 
	\end{figure}

	The above experimental results have already shown that DFF-ATMF can improve the performance of audio-text Sentiment analysis. We now analyze the attention values to understand the learning behavior of the proposed architecture better. 

We take a video from the CMU-MOSI test set as an example. From the attention heatmap in Figure \ref{Fig.main5},  we can see evidently that by applying different weights across contextual utterances and modalities, the model is able to predict labels of all the utterances correctly, which shows that our proposed fusion strategy with multi-feature and multi-modality is indeed effective, and thus has good feature complementarity and excellent robustness of generalization ability. However, at the same time, we have a doubt about the multi-feature fusion. When the raw waveform of the audio is fused with the vector of acoustic features, the dimensions are inconsistent. If the existing method is utilized to reduce the dimension, some audio information may also be lost. We intend to solve this problem from the perspective of some mathematical theory such as the angle between two vectors. 
	
	Similarly, the attention weight distribution heatmaps on the CMU-MOSEI and IEMOCAP test sets are shown in Figure \ref{Fig.main6} and \ref{Fig.main7}, respectively. Furthermore, we also give the softmax attention weight comparison of the CMU-MOSI, CMU-MOSEI, and IEMOCAP test sets in Figure \ref{Fig.main8}.
	
	\section{Conclusions}
	In this paper, we propose a novel fusion strategy, including multi-feature fusion and multi-modality fusion, and the learned features have strong complementarity and robustness, leading to the most advanced experimental results on the audio-text multimodal sentiment analysis tasks. Experiments on both the CMU-MOSI and CMU-MOSEI datasets show that our proposed model is very competitive. More surprisingly, the experiments on the IEMOCAP dataset achieve unexpected state-of-the-art results, indicating that DFF-ATMF can also be generalized for multimodal emotion recognition. In this paper, we did not consider the video modality because we try to use only the information of audio and text derived from videos. To the best of our knowledge, this is the first attempt in the multimodal domain. In the future, we will consider more fusion strategies supported by basic mathematical theories for multimodal sentiment analysis.
	
	\section{Acknowledgements}
	
	This research work was supported by the National Undergraduate Training Programs for Innovation and Entrepreneurship (Grant No. 201810022064) and the World-Class Discipline Construction and Characteristic Development Guidance Funds for Beijing Forestry University (Grant No. 2019XKJS0310). We also thank the anonymous reviewers for their thoughtful comments. Special thanks to the support of AAAI 2020 and AffCon2020.

\bibliography{references.bib}

\bibliographystyle{aaai}

\end{document}